\documentclass{article}

\usepackage[final]{corl_2017} 

\usepackage[numbers]{natbib}
\usepackage{multicol}
\usepackage{amsmath}
\usepackage{amsfonts}
\usepackage{amssymb}
\usepackage{booktabs}
\usepackage{multirow}
\usepackage{dblfloatfix}
\usepackage[ruled, noend, linesnumbered]{algorithm2e}
\SetAlFnt{\small}
\usepackage[]{graphicx}
\usepackage{subcaption}
\usepackage{enumitem}
\usepackage{wrapfig}

\newcommand{\ra}[1]{\renewcommand{\arraystretch}{#1}}
\DeclareMathOperator*{\argmin}{argmin}
\DeclareMathOperator*{\argmax}{argmax}

\title{Fastron: An Online Learning-Based Model and Active Learning Strategy for Proxy Collision Detection}

\author{
  Nikhil~Das, Naman~Gupta, and Michael~Yip\\
  Electrical and Computer Engineering\\
  University of California, San Diego
}

\begin{document}
\maketitle


\begin{abstract}
We introduce the Fastron, a configuration space (C-space) model to be used as a proxy to kinematic-based collision detection. The Fastron allows iterative updates to account for a changing environment through a combination of a novel formulation of the kernel perceptron learning algorithm and an active learning strategy. Our simulations on a 7 degree-of-freedom arm indicate that proxy collision checks may be performed at least 2 times faster than an efficient polyhedral collision checker and at least 8 times faster than an efficient high-precision collision checker. The Fastron model provides conservative collision status predictions by padding C-space obstacles, and proxy collision checking time does not scale poorly as the number of workspace obstacles increases. All results were achieved without GPU acceleration or parallel computing.
\end{abstract}

\keywords{configuration space, collision detection} 


\section{Introduction}
Configuration space (C-space) is a space that completely defines every kinematic configuration of the robot \citep{choset2005principles}. Robot configurations that are not in collision with workspace obstacles comprise the $C_{free}$ regions of C-space, and the $C_{obs}$ regions denote configurations in which the robot is in collision with a workspace obstacle. Checking for collisions is often a computational burden for robots working in environments with obstacles, but is a necessity for processes in which the robot must interact with or navigate through its environment, such as with Rapidly-Exploring Random Trees (RRTs) \citep{Lavalle98rrt}, a sampling-based motion planning algorithm.

A difficulty in working with C-space is that obstacle geometries generally do not trivially map from the workspace to C-space \citep{choset2005principles,hwang1992gross}. Sampling-based motion planners instead spend a large majority of their computation time on performing collision checks \citep{Elbanhawi} to infer C-space obstacles. In the case of workspaces with moving obstacles, $C_{obs}$ changes non-trivially, which makes maintenance of an updated map in C-space for collision detection a bottleneck in performance. Specialized hardware such as FPGAs \citep{Murray-RSS-16} accelerates the collision detection step, but algorithmic solutions may reduce the overall computation, which in turn may further improve hardware-based solutions.

\subsection{Contributions}
Realizing the high cost involved in kinematic-based collision detections (KCDs), we seek to decrease the computational cost of collision checking by learning a proxy collision detector that efficiently learns and maintains C-space representations that change over time. In this paper, we present the Fastron algorithm, a fast technique to generate and update an approximate C-space representation for proxy collision checking.

The purpose of these efforts is to reduce the computation required for collision checking for processes that suffer from a large number of KCDs so that more resources may be dedicated toward other computationally-intensive tasks, including further sampling for fine motion planning or model updates for reinforcement learning algorithms. Integrating the Fastron into motion planning algorithms is an obvious utilization, yet other highly-iterative applications that rely on collision detection could benefit from the Fastron, such as reward evaluation for reinforcement learning for simulated robot manipulation tasks and approximate object interactions in physics or CAD simulations.

A learning-based approach to modeling C-space is advantageous because a lightweight model and intelligent information gathering may be used in lieu of dense representation and sampling of a typically large-dimensional space. The Fastron is based on a modification of the kernel perceptron learning algorithm and uses a novel active learning strategy to reduce the total number of KCDs in favor of faster, proxy collision checks. Active learning algorithms select which samples to query so as to potentially reduce the number of queries to an oracle (who provides true labels at a higher cost) to perform during training or model updates \cite{manocha2013efficient,schohn2000less}. In the case of C-space estimation, active learning is useful when selecting on which samples accurate yet costly KCDs should be performed. The Fastron algorithm updates iteratively using periodic snapshots of obstacles' shapes and locations in the reachable workspace. Prior knowledge of all potential obstacle geometry models and trajectories is not required.

The novel contributions of this paper are:
\begin{enumerate}[topsep=0pt,itemsep=-1ex,partopsep=1ex,parsep=1ex]
\item a simple yet efficient method to learn and represent C-space obstacles using a kernel perceptron decision boundary
\item a modified kernel perceptron that allows both addition and removal of support points, and
\item an active learning strategy to efficiently search for collision status changes in a changing environment, where there is limited computation time between control cycles.
\end{enumerate}

\begin{figure}[b]
  \includegraphics[width=0.8\textwidth]{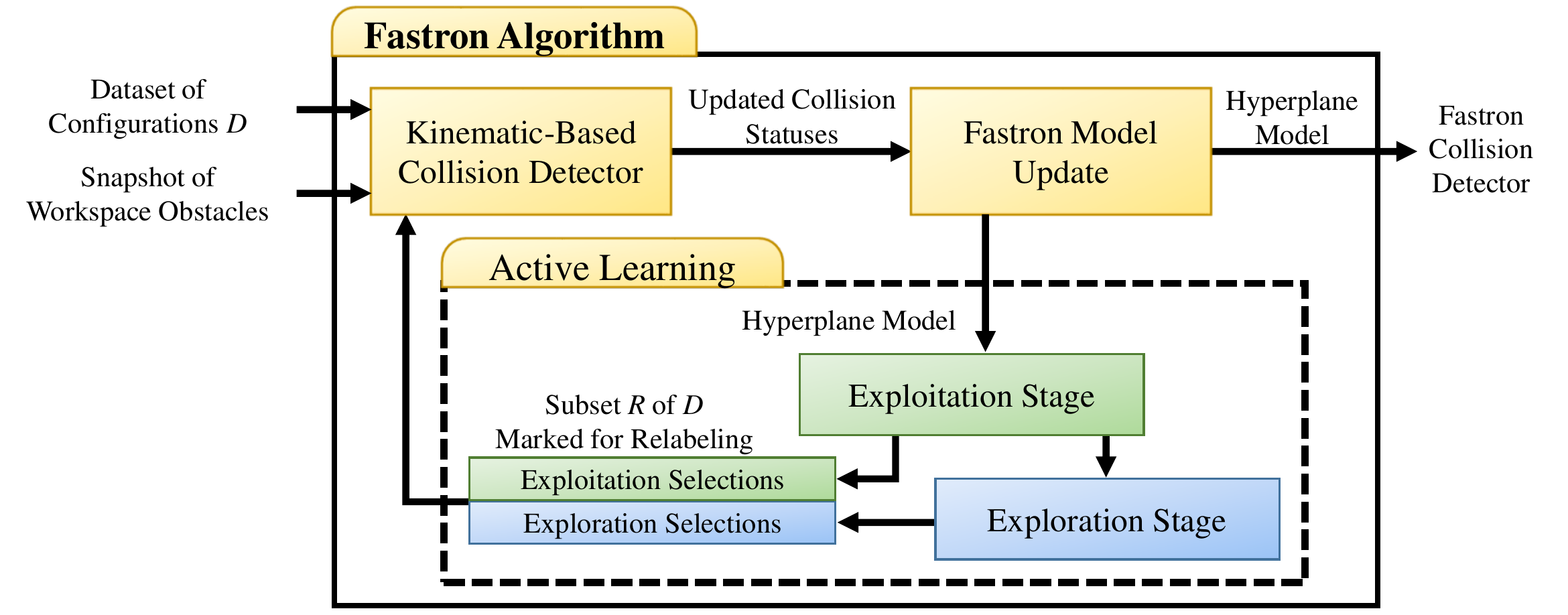}
  \caption{Pipeline of Fastron algorithm for generating and updating the C-space model used for fast collision checking.}
  \label{fig:pipeline}
\end{figure}

\subsection{Related Work}
As with our Fastron method, previous works have utilized machine learning-based models to approximate $C_{free}$ and $C_{obs}$ based on sampled configurations and use active learning strategies to guide the search for new information to update or refine the models. The following are contributions toward representing C-space environments with learning-based models.

\citet{manocha2013efficient} use an incremental support vector machine (SVM) to learn an accurate representation of C-space between two objects in an offline step. Their active learning strategy exploits the structure of the SVM-based hyperplane to add new points in order to construct a near-perfect representation of C-space obstacles. A new classifier must be precomputed for each pair of objects, thereby increasing the training time and proxy collision detection time. Additionally, since the models are learned in an offline stage, the geometry models of all workspace obstacles must be known a priori, which is not always a luxury. Online implementations would fare poorly when new obstacles are introduced into the workspace since this would require learning a completely new SVM model, which is unsuitable for real-time applications.

\citet{Huh} use Gaussian mixture models (GMM) to represent $C_{free}$ and $C_{obs}$, from which proxy collision detection is performed by assigning a query configuration the same label as the closest Gaussian. Their iterative GMM technique allows the model to update when workspace obstacles move to intersect a planned trajectory. A limitation of the GMM approach is the model may not fit irregularly-shaped $C_{obs}$ regions effectively as GMMs use a limited number of Gaussians. Additionally, the underlying generative models are updated to fit new information, which consequently does not guarantee the resulting discriminative classifier immediately fits the new information.

\citet{Burns-RSS-05} use a k-nearest neighbors (k-NN) model for C-space, which is not intended to be used in the case of moving workspace obstacles. \citet{pan2016} also use a k-NN model, accelerated by locality-sensitive hashing (LSH). Their method significantly reduces the time required for collision checking for sampling-based motion planners by building a database to use for k-NN queries. Though not implemented in a changing environment, they propose their method can extend to a changing environment by gridding the workspace and only performing collision checks on configurations associated with dynamic cells.

\section{Methods}
In this section, we provide a detailed description of the Fastron algorithm. The steps of the algorithm are summarized in the block diagram in Fig. \ref{fig:pipeline}. The algorithm cycles through two steps: updating the collision boundary model (\ref{modeling}) and active learning to search for collision status changes (\ref{activeLearning}).

\subsection{Modeling C-Space Using Perceptron}
\label{modeling}
We require (and the Fastron offers) a model that
\begin{enumerate}[topsep=0pt,itemsep=-1ex,partopsep=1ex,parsep=1ex]
\item is fast to train, 
\item is fast in classifying query configurations, 
\item adequately fits training data,
\item attempts to reduce mistakes where $C_{obs}$ configurations are classified as $C_{free}$,
\item has an easily exploitable structure to facilitate the search for collision status changes, and 
\item can efficiently account for collision status changes without retraining from scratch.
\end{enumerate}

The batch kernel perceptron algorithm, which identifies a set of support points defining a separating hyperplane between two classes, satisfies the first three requirements and thus serves as the base model for the algorithm. We modify the kernel perceptron to satisfy the remaining requirements. This section describes the original batch kernel perceptron algorithm and our modifications. Pseudocode for the modified perceptron algorithm is shown in Algorithm \ref{alg:perceptron}.


\begin{algorithm}[b]
\caption{Fastron Model Updating}
\label{alg:perceptron}
 \KwIn{Weight vector $\alpha$; hypothesis vector $F$; Gram matrix $G$ for a dataset $\mathcal{D}$; true labels $y$ for $\mathcal{D}$; conditional bias parameter $r^+$; maximum number of updates $maxUpdates$}
 \KwOut{Updated $\alpha$; updated $F$}

 \For {$iter = 1$ to $maxUpdates$}
 {
 \tcp{Remove redundant support points}
     \While{$\exists$ i s.t. $y_i (F_i - \alpha_i) > 0$ and $\alpha_i \neq 0$}{
			$j \leftarrow \argmax_i y_i(F_i - \alpha_i)$ \\
            $F_i \leftarrow F_i - G_{ij}\alpha_{j} \, \forall i$\\
            $\alpha_j \leftarrow 0$
        }
 \tcp{Margin-based prioritization}
 \uIf{$ y_i F_i > 0$ $\forall i$}{
       \Return{$\alpha, F$}
   }
  \uElse{
  $j \leftarrow \argmin_i y_i F_i$
  }
  \tcp{One-step weight correction with conditional biasing}
 	\uIf{$y_j > 0$}{
    $\Delta\alpha \leftarrow r^+y_j-F_j$
    }
    \uElse{
    $\Delta\alpha \leftarrow y_j-F_j$
    }
    
    $\alpha_j \leftarrow \alpha_j + \Delta \alpha$ \\
    $F_i \leftarrow F_i + G_{ij}\Delta\alpha \, \forall i$
 }
 \Return{$\alpha, F$}
\end{algorithm}

\begin{figure}[b]
	\centering
    \begin{subfigure}[b]{0.4\textwidth}
        \includegraphics[width=\textwidth]{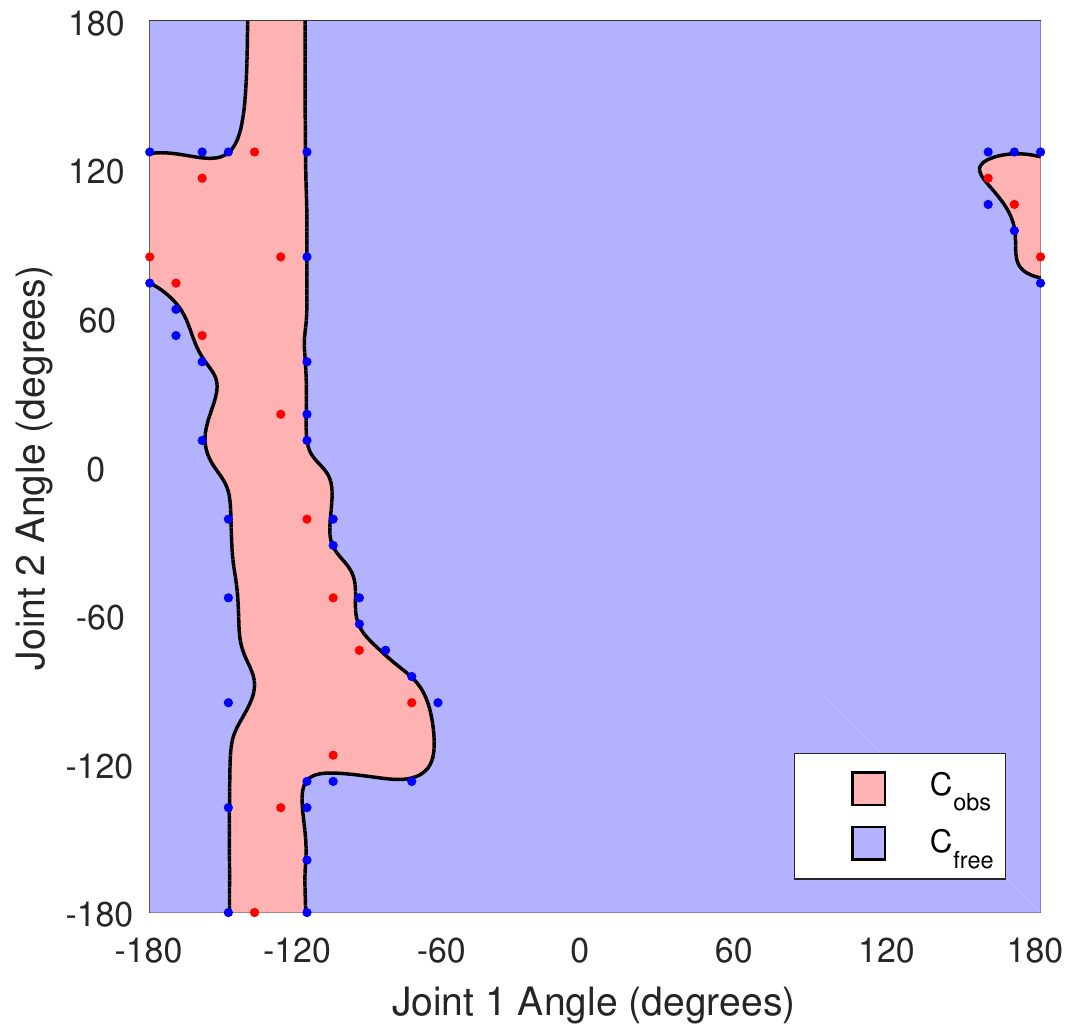}
		\caption{}
        \label{fig:newCspace}
    \end{subfigure}
    \begin{subfigure}[b]{0.4\textwidth}
        \includegraphics[width=\textwidth]{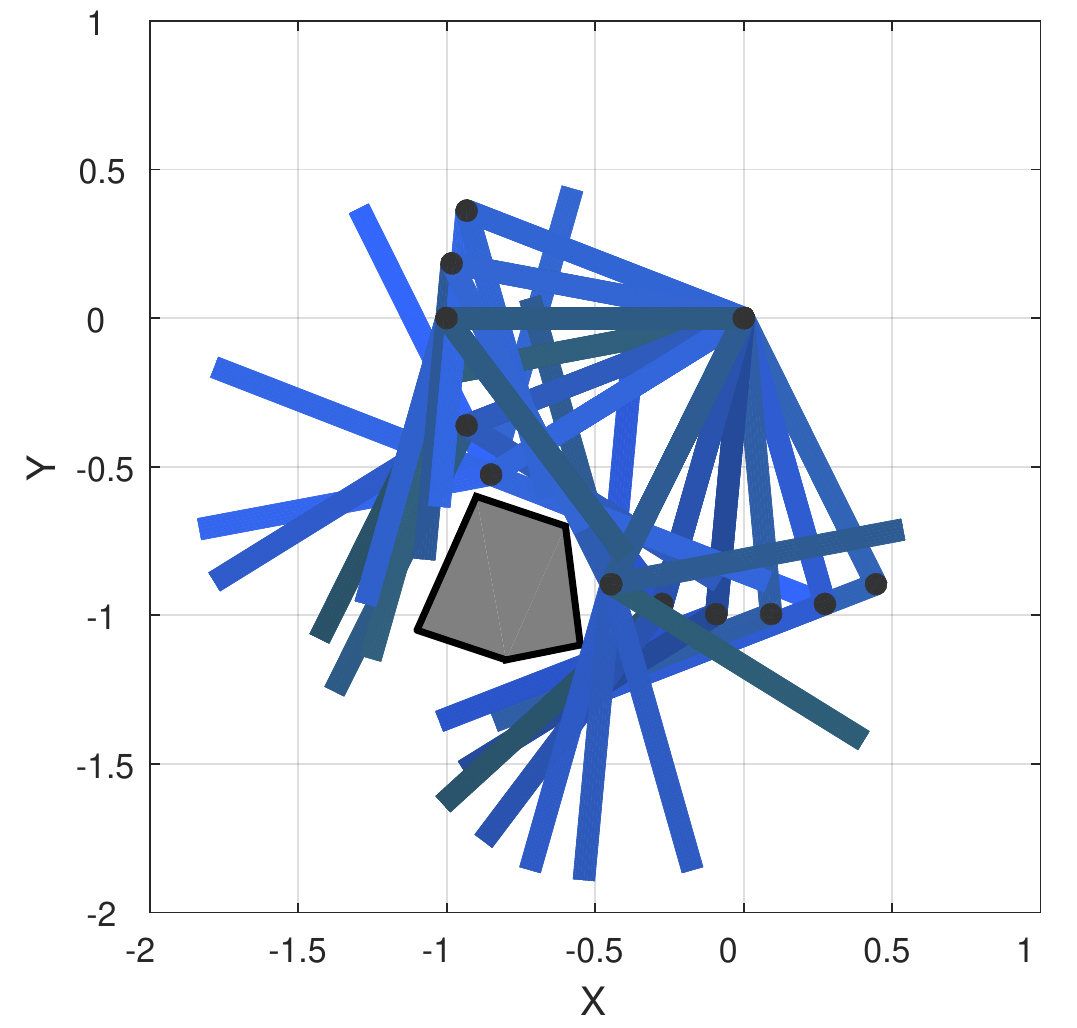}
        \caption{}
        \label{fig:newCfreeConfigs}
    \end{subfigure}
      \caption{(a) Decision boundary (black curve) and support points (red and blue points) learned by our modified kernel perceptron. (b) Workspace representations of $C_{free}$ support points from our modified kernel perceptron (blue 2 DOF manipulators) and a workspace obstacle (gray polygon).}
  \label{fig:oldVsNewPerceptron}
\end{figure}

\subsubsection{Training and Classification with Original Kernel Perceptron}
The original batch kernel perceptron algorithm trains a model that may be used to classify a query point into one of two classes. During training, the model updates when it encounters a training point that it would misclassify. Given a training dataset $\mathcal{D}$ of $N$ labeled samples, the kernel perceptron algorithm learns a hypothesis $f(x)$, which has the form $\sum_i{\alpha_iK(x_i,x)}$, where $\alpha \in \mathbb R ^N$ is a sparse weight vector, $K(\cdot,\cdot)$ is the Gaussian kernel function, and $x_i$ is a sample in $\mathcal{D}$ with a known label $y_i\in [-1,+1]$. The Gaussian kernel is defined as $K(x_i, x_j) = exp(-\gamma \|x_i-x_j\|^2)$, where $\gamma$ is a parameter specifying the narrowness of the Gaussian. The goal of the perceptron algorithm is to define $\alpha_i$ such that the margin $y_if(x_i)$ for each training point $x_i$ is positive. The original algorithm learns $\alpha_i$ by shuffling $\mathcal{D}$ and computing $y_if(x_i)$ for each $x_i$. Whenever $y_i f(x_i)\leq 0$, $y_i$ is added to $\alpha_i$. This shuffle and update procedure is repeated until all training points have a positive margin or an epoch limit has been reached.

The hypothesis at each sample can be written in vector form as $F=G\alpha$, where the $i^{th}$ element of $F$ is $f(x_i)$ and $G$ is the kernel Gram matrix for the $N$ datapoints. To avoid redundant matrix-vector multiplications, we can store $F$ and add (or subtract) the $i^{th}$ column of $G$ whenever we increment (or decrement) $\alpha_i$. The update rule for the original kernel perceptron may thus be written as 
\begin{gather}
\alpha_i \leftarrow \alpha_i + y_i \\
F \leftarrow F + y_iG_{*i}
\end{gather}
where $G_{*i}$ is the $i^{th}$ column of $G$.

The support set $S$ is the set of points in $\mathcal{D}$ with a nonzero weight in $\alpha$. The support points that comprise $S$ may be used to classify a query configuration $x$ as $\hat{y} (x) = sgn\left(\sum_{i:x_i\in S}{\alpha_i K(x_i, x)}\right)$. We may use this classification as a proxy collision check where $\hat{y} = \pm  1$ represents an in-collision or a collision-free status, respectively.

\subsubsection{One-Step Weight Correction and Conditional Biasing}
The original kernel perceptron algorithm increases the weight of a misclassified point $x_i$ by $y_i$, but $x_i$ will still be incorrectly classified if the magnitude of the margin $\|y_i f(x_i)\|$ prior to update is greater than 1. The appropriate value to assign to weight $\alpha_i$ to ensure $x_i$ is correctly classified may be easily realized based on the requirement that the resulting margin must be positive. It is evident that for $x_i$ to be classified correctly, $\alpha_i$ must equal $ry_i-\sum_{j\neq i}{\alpha_j K(x_j,x_i)}$, where $r>0$. We can avoid computing the summation in the second term by noting the change in $\alpha_i$ after the update is $\Delta\alpha_i=ry_i-\sum_{j}{\alpha_j K(x_j,x_i)}=ry_i-f(x_i)$. Thus, the update rule for our modified perceptron is:
\begin{gather}
\alpha_i \leftarrow \alpha_i + \Delta\alpha_i \\
F \leftarrow F + \Delta\alpha_i G_{*i}
\end{gather}

The advantage of this modification is the misclassified point $x_i$ is guaranteed to be modeled correctly after the update. To increase the safety of the hyperplane, we conditionally set $r$ depending on the label of the support point we are adding to $S$. More explicitly, we define a conditional bias parameter $r^+>1$, and we set $r=r^+$ when $y_i>0$ and $r=1$ when $y_i<0$. When $r^+$ is greater than 1, $C_{obs}$ configurations have a larger influence on the update to the hyperplane compared to $C_{free}$ configurations which slightly pads the C-space obstacles, thereby potentially reducing the false negatives (misclassification of a $C_{obs}$ configuration as $C_{free}$).
    
\subsubsection{Margin-Based Prioritization}
The magnitude of a point's margin indicates how confidently the point is assigned to its predicted label. By updating the weight associated with the most negative margin, the most erroneous point is forced to be correctly classified using the one-step weight adjustment described above. Thus, rather than shuffling the data and running through $\mathcal{D}$ in a random order, we choose to update $\alpha_i$ where $i=\argmin_j⁡{y_j f(x_j)}$. The advantage of margin-based prioritization is that the support points end up closer to the decision boundary. Placing support point near the boundary grants the ability to exploit the structure of the model to search for collision status changes near the boundary. 

\subsubsection{Redundant Support Point Removal}
A support point should be removed from $S$ (but remain in $\mathcal{D}$) when it is redundant. Redundant support points are those that will be correctly classified even if their corresponding $\alpha$ value is 0, i.e., $\{x_i|x_i \in S \wedge y_i (F_i - \alpha_i) > 0\}$. Support points are removed in decreasing order of positive resultant margin by setting the weight to 0 and updating $F$ accordingly. The removal step is complete once $y_i (f(x_i)-\alpha_i )<0\,\forall i$, i.e., removing another support point will cause it to be misclassified.

	Redundant support point removal is useful when the collision status of the points in $\mathcal{D}$ change in response to a dynamic environment, causing the updated hyperplane to shift away from previous support points. Removing redundant support points ensures that the support points are as close as possible to the hyperplane. Additionally, without redundant support point removal, it is possible that eventually $S = D$, which slows classification performance by forfeiting the sparsity of the model.

\begin{algorithm}[t]
\caption{Fastron Active Learning Strategy}
\label{alg:activeLearning}
 \KwIn{KCD allowance $A$; exploitation proportion $p$; support set $S$; dataset $\mathcal{D}$; Gram matrix $G$; maximum number of nearest non-support points $k_{NS}$}
 \KwOut{Set of points $R\subset \mathcal{D}$ to be relabeled}


\tcp{Exploitation Stage}
\uIf{$|S|\leq A$}{
  $R\leftarrow  S$ \\
  \For{$k = 1$ to $ k_{NS}$}{
    \uIf{$|R|<pA$}{
    	$R \leftarrow R \cup knnsearch(\mathcal{D}\backslash S, S, k)$ \label{alg:knnsearch}
    }
  }
}
\uElse{
	$R\leftarrow sample(S,A)$
}
\tcp{Exploration Stage}
$R\leftarrow R \cup sample(\mathcal{D}\backslash R,A-|R|)$

 \Return{$R$}
\end{algorithm}

\subsection{Active Learning for Efficient Relabeling} \label{activeLearning}
In response to a changing environment, the collision statuses of the points in $\mathcal{D}$ must be updated before updating the hyperplane model. To know with absolute certainty which points have switched labels, KCD must be performed on each point in $\mathcal{D}$, which clearly is a time-consuming and potentially unnecessary process. Instead, the Fastron selects a subset $R$ of $\mathcal{D}$ to relabel, where the maximum value of $|R|$ is set by a user-defined allowance $A$ for the total number of KCDs to perform per model update.

Points are selected to be included in $R$ using a two-stage active learning strategy. A common active learning strategy is to balance exploitation of the current model and exploration of the entire space, which is the technique the SVM C-space approach uses \citep{manocha2013efficient}. The Fastron adopts a similar active learning strategy, but in the interest of efficient model updating, the Fastron selects samples in $\mathcal{D}$ to relabel rather than generating entirely new samples. This allows the Fastron to take advantage of precomputed distances between points rather than recomputing distances.

\begin{wrapfigure}[20]{r}{0.5\textwidth}
	\centering
	\includegraphics[width=0.42\columnwidth]{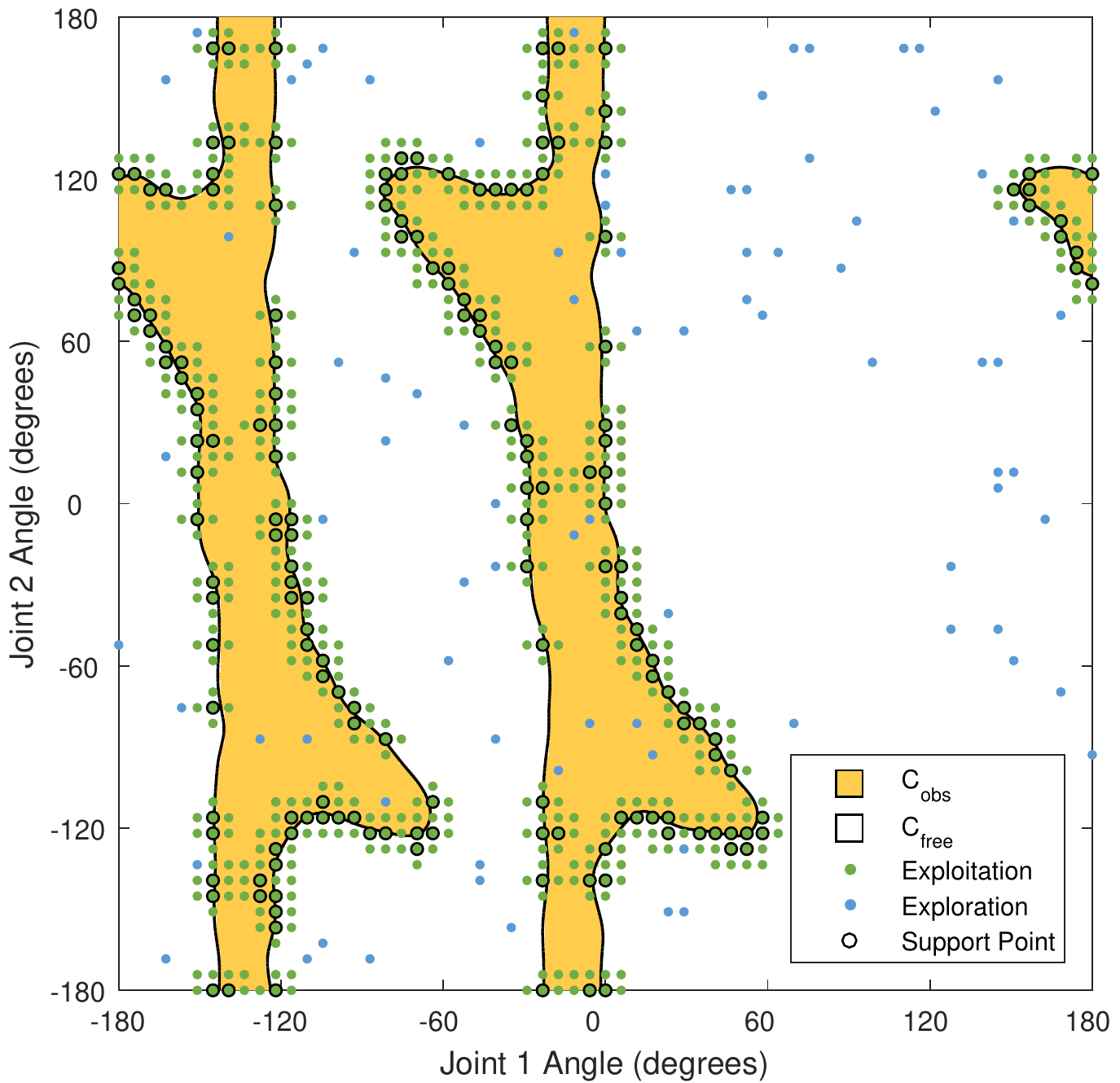}
	\caption{Example set of samples $R$ selected by the active learning strategy for relabeling via KCD.}
	\label{fig:activeLearningExample}
\end{wrapfigure}

The strategy marks at least $pA$ points in the exploitation stage, where $p$ is a user-defined proportion of the allowance dedicated for exploitation. The remainder of the allowance is exhausted in the exploration stage. The two stages are described in the following subsections. Pseudocode for our active learning strategy is provided in Algorithm \ref{alg:activeLearning}, and an example set of points selected by the strategy is shown in Fig. \ref{fig:activeLearningExample}.
\subsubsection{Exploitation Stage}
Assuming that movements of the workspace obstacles cause small perturbations of the corresponding C-space obstacles, the Fastron first checks for status changes near the boundary of the C-space obstacles. This is accomplished by exploiting the structure of the perceptron model, which typically has its support points near the decision boundary when using our modified perceptron.
At the beginning of each model update, $R$ is initialized to the empty set. All current support points are then included in $R$. In the case that including all support points will exceed the allowance $A$, $A$ support points are randomly chosen to be included in $R$. After adding support points to $R$, if $|R|$ is less than $pA$, each support point's $i^{th}$-nearest non-support point is iteratively included until either the resulting $|R|$ is greater than or equal to $pA$ or $k_{NS}|S|$ non-support points have been included in $R$, where $k_{NS}$ is a user-defined amount.

Distance information between points is conveniently available in Gram matrix $G$, and since the values of $G$ do not change throughout the lifetime of the Fastron algorithm, costly conventional k-NN searches or more efficient approximations are not necessary. Line \ref{alg:knnsearch} in Algorithm \ref{alg:activeLearning} assumes the k-NN search utilizes the distance information in $G$.

\subsubsection{Exploration Stage}
If we have not yet exhausted the collision check allowance, the remainder of the allowance is utilized by randomly selecting configurations. $A - |R|$ points are randomly selected from $\mathcal{D}\setminus R$. The purpose of this random exploration step is to search for new or drastically different C-space obstacles, such as when a new object enters the reachable workspace or an existing object moves quickly.
	

\section{Experimental Results}

\subsection{Experiments on 2 DOF Manipulator}
We perform preliminary experiments on a 2 DOF manipulator to easily visualize both the workspace and C-space. We create random convex polygonal workspace obstacles, and use the Gilbert-Johnson-Keerthi (GJK) algorithm \cite{GJK} for KCDs. We perform all 2 DOF simulations in MATLAB without the use of GPU acceleration or parallel computing to demonstrate its native speed.
    
We compare the collision detection time of FCDs and KCDs under increasingly difficult conditions (increasing number of obstacles in the workspace). We use $N = 625$, kernel width $\gamma = 10$, and conditional bias parameter $r^+ = 100$ for our Fastron model parameters, where $\gamma$ and $r^+$ were selected via cross-validation. In the interest of generating a safe model, recall (true positive rate, or percentage of $C_{obs}$ configurations correctly classified) is our primary metric for performance. High values of recall indicate that the model rarely considers $C_{obs}$ configurations to be in $C_{free}$. Table \ref{table:numObstacles2} demonstrates the performance of FCD for various numbers of workspace obstacles. Recall remains high (over 98\%) as the number of obstacles increases. Table \ref{table:numObstacles2} also includes false positive rate (FPR) to demonstrate the effect of padding due to conditional biasing in a more crowded workspace. FPR increases along with the number of obstacles because the Fastron has a bias toward labeling configurations as $C_{obs}$ in regions of uncertainty. The speed improvement of FCD over KCD drastically increases as the number of workspace obstacles increases, showing that FCDs are more resilient to the number of obstacles than KCDs.

\begin{table}
\scriptsize
\centering
\ra{1.3}
\begin{tabular}{@{}lccccc@{}}
\multicolumn{1}{c}{}& \multicolumn{5}{c}{Number of Obstacles} \\
\cmidrule{2-6}
\multicolumn{1}{c}{}& 1&2&3&4&5 \\
\cmidrule{2-6}
\multicolumn{1}{l}{FCD Recall (\%)}&98.3& 	98.3& 	98.5	& 98.9& 	98.9 \\
\multicolumn{1}{l}{FCD FPR (\%)}&3.6& 	6.7& 	11.5	& 13.9& 	16.0 \\
\multicolumn{1}{l}{FCD Time ($\mu s$)}&33.8	& 37.9& 	39.2	& 39.6& 	40.5 \\
\multicolumn{1}{l}{\textbf{Ratio of KCD to FCD Time}} & \textbf{4.9}	& \textbf{7.5}	& \textbf{9.4}& 	\textbf{11.1}& 	\textbf{12.0} \\
\end{tabular}
\caption{Recall, false positive rate, and collision check time of FCDs for 2 DOF manipulator with various number of obstacles. Note that KCD timings scale poorly with the number of obstacles, while FCDs do not.}
\label{table:numObstacles2}
\end{table}

\begin{table}[t]
\scriptsize
\centering
\begin{tabular}{@{}lccrcccrcccr@{}}
\multicolumn{1}{c}{}& \multicolumn{3}{c}{$A = 0.1N$} & \multicolumn{1}{c}{}& \multicolumn{3}{c}{$A = 0.3N$} & \multicolumn{1}{c}{}& \multicolumn{3}{c}{$A = 0.5N$} \\
\multicolumn{1}{c}{}& Recall & FPR & Time && Recall & FPR & Time && Recall & FPR & Time\\
\cmidrule{2-12}
$N=100 $& 75.0 & 6.5 & 1.5 && 84.2 & 7.5 & 3.4 && 85.5 & 7.8 & 4.7 \\
$N=400 $& 94.6 & 2.7 & 5.6 && 95.4 & 3.3 & 12.6 && 93.9 & 2.9 & 16.6 \\
$N=625 $& 91.0 & 2.0 & 8.3 && 95.4 & 2.2 & 18.8 && 95.7 & 2.0 & 26.2 \\
$N=900 $& 94.5 & 1.6 & 13.2 && 96.5 & 1.6 & 26.4 && 93.8 & 1.4 & 36.3 \\
$N=1225 $& 95.6 & 1.3 & 17.0 && 95.9 & 1.4 & 37.5 && 95.6 & 1.0 & 48.4 \\
\end{tabular}
\caption{Recall (\%), false positive rate (\%), and update time ($ms$) for various dataset sizes $N$ and exploitation stage proportions $p$ for 2 DOF manipulator in a changing environment.}
\label{table:accRecallUpdateTime2}
\end{table}

\begin{table}
\scriptsize
\centering
\begin{tabular}{@{}llccc@{}}
\multicolumn{2}{c}{}& \multicolumn{3}{c}{Number of Obstacles} \\
\cmidrule{3-5}
\multicolumn{2}{c}{}& 1&2&3 \\
\cmidrule{3-5}

\multicolumn{1}{c}{FCL}&\multicolumn{1}{l}{FCD Recall (\%)}& 92.8 &  95.3 &  98.1 \\
\multicolumn{1}{c}{}&\multicolumn{1}{l}{FCD FPR (\%)}&14.3 & 22.9  & 	30.9 \\
\multicolumn{1}{c}{}&\multicolumn{1}{l}{FCD Time ($\mu s$)}& 4.1 &  4.0 & 	 4.2 \\
\multicolumn{1}{c}{}&\multicolumn{1}{l}{\textbf{Ratio of KCD to FCD Time}} & \textbf{8.1} &  \textbf{9.4}	& \textbf{10.3}\\
\cmidrule{1-5}
\multicolumn{1}{c}{GJK}&\multicolumn{1}{l}{FCD Recall (\%)}&91.6 & 	94.0  & 96.0 \\
\multicolumn{1}{c}{}&\multicolumn{1}{l}{FCD FPR (\%)}&7.2 & 	11.1 & 	32.6 \\
\multicolumn{1}{c}{}&\multicolumn{1}{l}{FCD Time ($\mu s$)}&3.6  & 4.0  & 	4.6 \\
\multicolumn{1}{c}{}&\multicolumn{1}{l}{\textbf{Ratio of KCD to FCD Time}} & \textbf{2.0}  & \textbf{2.7} 	& \textbf{2.9} \\

\end{tabular}
\caption{Recall, false positive rate, and collision check time of FCDs for 7 DOF manipulator with various number of obstacles. KCD timings scale poorly with the number of obstacles, while FCDs do not.}
\label{table:numObstacles7}
\end{table}

	We evaluate the performance of the Fastron in an environment with a moving randomly-generated polygon under various dataset sizes $N$ and relabeling allowances $A$, with $\gamma = 10$, exploitation proportion $p=0.8$, and a maximum nearest non-support point number $k_{NS}=4$. We tabulate the average recall, FPR, and update time (model updating and active learning) over 10 second trials in Table \ref{table:accRecallUpdateTime2}. Compared to the static case shown in \ref{table:numObstacles2}, recall is lower in the moving obstacle case possibly because all collision status changes may not have been detected. However, recall is still large (over 90\%) for $N$ larger than or equal to 400. Update time worsens as $p$ increases because more KCDs are required. FPR decreases for increasing $N$ because when there are more points distributed in C-space, there is a decreased requirement for the Fastron to be conservative by padding C-space obstacles in regions of uncertainty.


\begin{table}[t]
\scriptsize
\centering
\begin{tabular}{@{}llccrcccrcccr@{}}
\multicolumn{2}{c}{}& \multicolumn{3}{c}{$A = 0.1N$} & \multicolumn{1}{c}{}& \multicolumn{3}{c}{$A = 0.3N$} & \multicolumn{1}{c}{}& \multicolumn{3}{c}{$A = 0.5N$} \\
\multicolumn{2}{c}{}& Recall & FPR & Time && Recall & FPR & Time && Recall & FPR & Time\\
\cmidrule{3-13}
FCL&$N=1000 $& 98.9 & 36.0 & 2.7 && 98.9 & 38.2 & 2.9 && 98.8 & 37.5 & 3.1 \\
&$N=4000 $& 96.1 & 18.4 & 29.1 && 95.7 & 17.3 & 31.7 && 94.7 & 15.6 & 32.8 \\
&$N=8000 $& 90.2 & 8.5 & 116.5 && 90.2 & 8.2 & 130.6 && 87.8 & 6.7 & 138.1 \\
\cmidrule{1-13}
GJK&$N=1000 $& 95.2 & 16.7 & 2.2 && 94.5 & 14.9 & 2.3 && 94.1 & 14.5 & 2.4 \\
&$N=4000 $& 93.4 & 9.5 & 29.3 && 92.0 & 7.6 & 30.8 && 91.0 & 7.0 & 31.8 \\
&$N=8000 $& 93.1 & 7.7 & 123.3 && 91.7 & 5.9 & 131.5 && 90.6 & 5.1 & 138.4 \\
\end{tabular}
\caption{Recall (\%), false positive rate (\%), and update time ($ms$) for various dataset sizes $N$ and exploitation stage proportions $p$ for 7 DOF PR2 manipulator.}
\label{table:NvsAt7DOF}
\end{table}



\subsection{Experiments on 7 DOF Manipulator}
We apply the Fastron algorithm to a simulated 7 DOF PR2 arm in a C++ ROS environment with shape primitives as workspace obstacles. KCD is performed using both the Flexible Collision Library (FCL) \cite{FCL} collision checker and GJK in the Bullet physics library. In the FCL cases, the actual PR2 arm mesh is used. While FCL may be used for high-precision collision checking, it is a popular collision checking framework and starts with a broad phase collision check which makes many collision checks fast. In the GJK cases, the arm is simplified as a set of oriented bounding boxes to provide an instance of a high-speed but low-fidelity collision checking framework. We do not rely on GPU acceleration or parallelization to speed up any part of the algorithm. In all following simulations, we use a fixed value of kernel width $\gamma = 10$ and conditional bias parameter $r^+ = 2$.

With a dataset of size $N = 4000$, the recall is sufficiently large (over 90\%) with both FCL and GJK KCDs as shown in Table \ref{table:numObstacles7}. FPR increases as the number of obstacles increases because the C-space is high dimensional so the spacing of the 4000 points cause the Fastron to pad the C-space obstacles more. The speed improvement of the FCD over KCD increases as the number of obstacles increases.

We evaluate the performance of the model in changing environments under various dataset sizes $N$ and relabeling allowances $A$ by considering the average recall, false positive rate, and update time with exploitation proportion $p=0.5$ and a maximum nearest non-support point number $k_{NS}=4$. Table \ref{table:NvsAt7DOF} shows that update time increases with $A$ because active learning involves KCDs. Recall decreases as $N$ increases but is generally above 90\%, and false positive rate improves as $N$ increases.
 
We demonstrate one use case of the Fastron algorithm by implementing a standard RRT motion planner \cite{Lavalle98rrt} using FCDs and KCDs for collision checks, henceforth referred to as FCD-RRT and KCD-RRT, respectively. We choose the standard RRT due to its simplicity, yet we note that dynamic RRTs and other variants handling moving obstacles will see similar benefits from the Fastron. We repeatedly compute an RRT from scratch over the course of a 10 second trial, translating the workspace obstacle between each RRT plan to simulate a changing environment. The position of the workspace obstacle is randomly generated such that the arm cannot take a straight approach to the goal configuration. We employ model updates and active learning to ensure the Fastron accounts for the changing environment. We use $N = 4000$ and $A = 0.3N$ for the RRT experiments.

When using FCL for KCDs, the average time spent in the collision checking stage of the FCD-RRTs is 108 ms, while 399 ms is required for the KCD-RRT's collision checking stage. When using GJK for KCDs, the collision checking stage takes 104 ms for FCD-RRTs and 164 ms for KCD-RRTs. The times required for model updating and active learning (which together take around 30 ms) are included in the timings for the FCD-RRTs' collision checking stages. As the collision checking stage is 3.7 times faster in the high-precision FCL case and 1.6 times faster in the low-fidelity GJK case, the Fastron demonstrates that the collision check bottleneck sampling-based motion planners often face may be lessened, especially if KCDs needed for information gathering are parallelized.


\section{Concluding Remarks}
We present the Fastron algorithm as a method to quickly represent and update a learning-based C-space model to be used for fast, proxy collision detection. We note that the Fastron complements, but not entirely supplants, kinematic-based collision checks because KCDs still serve as an oracle for acquiring updated information during active learning. The advantage of utilizing a learning-based model to represent C-space is a dense representation of C-space is not required. Instead, only a few support points represent the decision boundary between $C_{free}$ and $C_{obs}$, whose structure may be exploited to reduce costly query evaluations of the oracle KCD function.
 
In future work, we will determine a method to incorporate resampling (rather than relabeling) to increase model precision and a method to provide a confidence score on the classification output to facilitate active learning by guiding the information search toward regions of low confidence.



\clearpage


\bibliography{example}  

\end{document}